\documentclass[letterpaper, 10pt, conference]{ieeeconf}  %
\IEEEoverridecommandlockouts   %
\usepackage{xcolor}
\usepackage{xspace}
\usepackage{booktabs}
\usepackage{graphicx}
\let\labelindent\relax
\usepackage{enumitem}
\usepackage{amsmath}
\usepackage{amssymb}
\usepackage{bm}
\usepackage[normalem]{ulem}
\usepackage{stfloats}          %
\usepackage[hidelinks,breaklinks]{hyperref}   %
\usepackage{caption}
\definecolor{figgreen}{HTML}{2CA02C}
\definecolor{figblue}{HTML}{1F77B4}
\definecolor{figorange}{HTML}{FF7F0E}

\newcommand{\std}[1]{{\scriptsize$\pm$#1}}

\newcommand{\etal}[0]{{\em et al.~}}
\newcommand{\eg}[0]{{\em e.g.,~}}
\newcommand{\ie}[0]{{\em i.e.,~}}

\DeclareRobustCommand{\method}{\rotatebox[origin=c]{180}{A}\textsc{gp}\xspace}

\newcommand{\website}[1]{%
  \begin{center}\small
  \textbf{Project page:}\ #1
  \end{center}}

\title{\LARGE \bf
An Empirical Study on What Matters for Viewpoint-Generalizable Policies in Visual Imitation Learning
}

\author{Mino Nakura$^{1*}$, Sriram Krishna$^{1*}$, Yufei Wang$^{1}$,
      Shubham Tulsiani$^{1}$, Zackory Erickson$^{1\dagger}$, David Held$^{1\dagger}$%
\thanks{$^{*}$Equal contribution. $^{\dagger}$Equal advising.}%
\thanks{$^{1}$Robotics Institute, Carnegie Mellon University, Pittsburgh, PA 15213, USA.}%
}

\begin{document}

\maketitle
\thispagestyle{empty}
\pagestyle{empty}

\begin{abstract}

Visual imitation learning is a promising approach to training robot manipulation policies capable of completing a wide variety of tasks. However, policies today remain brittle to viewpoint perturbations, making deployment in diverse environments a challenge. We present a controlled empirical study of which design choices allow visuomotor policies to generalize across viewpoints. We find that viewpoint generalization improves when dense visual tokens are retained and the action head participates in geometric reasoning. On a suite of simulated tasks that span a wide range of camera poses, we show that these design choices yield a policy that remains performant across viewpoints. 
As a practical consequence, a policy trained with these design choices also transfers zero-shot from simulation to the real world under random camera configurations.

\end{abstract}

\website{\nolinkurl{https://vgp-sim2real.github.io/}}

\begin{figure*}[!t]
  \centering
  \includegraphics[width=\textwidth]{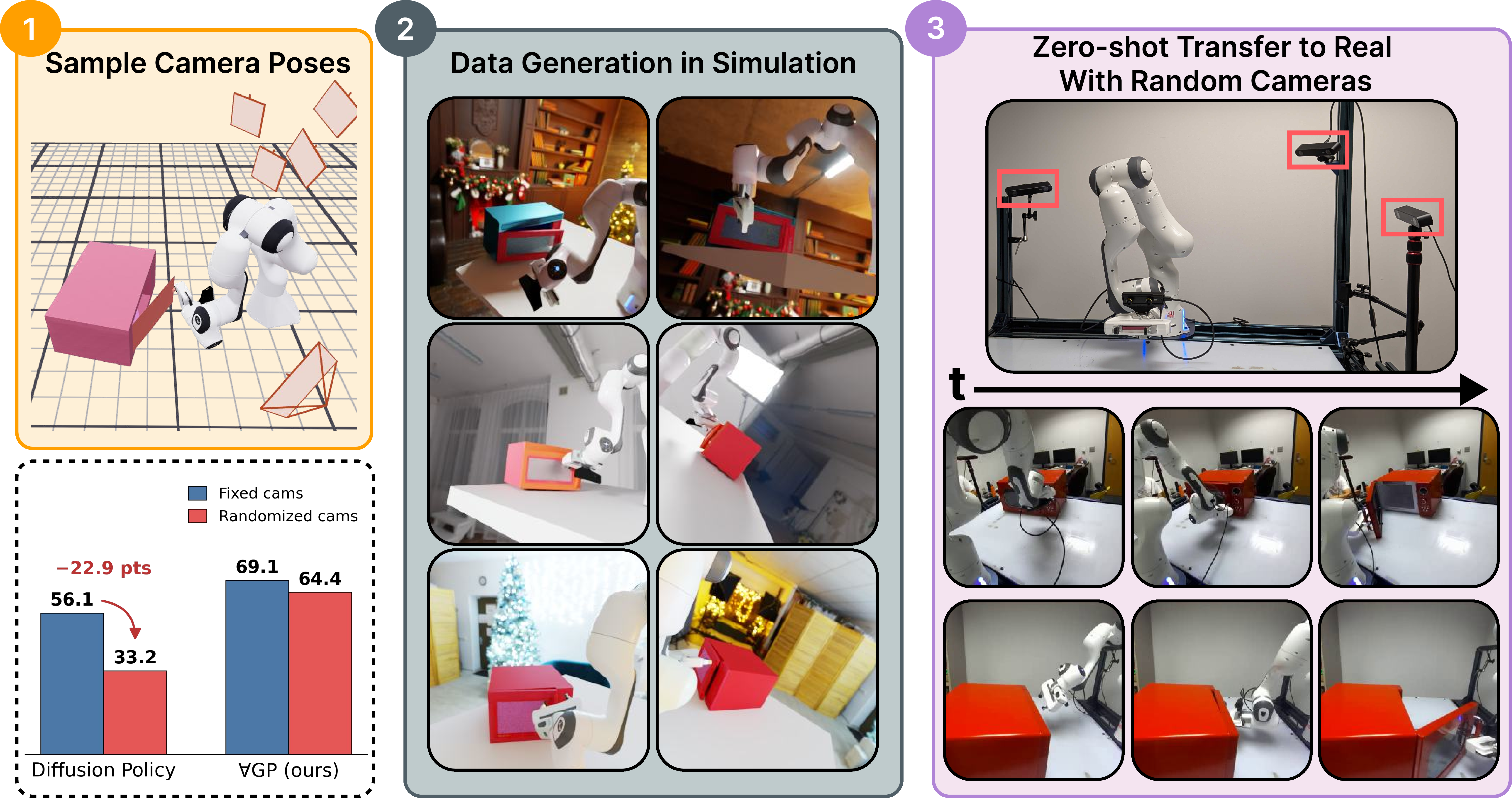}
  \caption{\textbf{Top Left:} We sample a wide distribution of camera poses in simulation. \textbf{Middle:} We generate demonstrations under these randomized views and train a policy that grounds visual, proprioception, and action tokens in a shared robot frame. \textbf{Right:} The policy transfers zero-shot to a real robot under random camera configurations.}
  \label{fig:teaser}
\end{figure*}

\section{Introduction}

A common approach to robot learning today is to train visuomotor policies via imitation learning~\cite{chi2023diffusionpolicy, zhao2023learning, Ze2024DP3}.
However, end-to-end policies that map pixels to actions inherit no structure about the robot's relationship to its observations, and must instead learn to localize the robot implicitly from the demonstration data. As a consequence, learned policies overfit to the camera configurations they were trained on and degrade sharply when the observation viewpoint shifts at deployment. This is particularly acute in settings where the robot is out-of-frame, leaving the policy with no visual anchor to localize the robot at all.
 
A growing body of work studies \textit{viewpoint-generalizable} manipulation policies. Policies that operate directly on 3D representations such as point clouds or voxels~\cite{Ze2024DP3, Wang2025articubot} are naturally robust to viewpoint shifts, since their inputs can be expressed in a scene-centric frame (\eg the robot base frame) rather than one tied to a particular camera. However, 3D visual encoders lack the large-scale pretraining available for image backbones, and existing 3D architectures do not scale efficiently to dense observations, requiring downsampling to execute at reasonable control frequencies. Image-based policies lack this natural viewpoint robustness, but recent work has attempted to narrow the gap by conditioning on explicit camera geometry~\cite{jiang2026knowyourcamera, kleeraven}. In contrast, we hypothesize that a policy does not need to explicitly know the location of the camera; rather, it needs to know the location of the pixels that the camera observes relative to the robot.
 
In this paper, we perform a study to identify the ingredients needed for viewpoint generalization. First, we find a large benefit in preserving dense visual features instead of relying on popular compression techniques~\cite{chi2023diffusionpolicy, Ze2024DP3} that reduce the scene information into a sparse global representation. Second, we show the benefits of incorporating recent advances in multi-view processing by building on the architecture of a visual geometry foundation model~\cite{lin2025depth}. Third, we observe that grounding every vision, proprioception, and action token in the robot frame via rotary positional encodings~\cite{su2021roformer} substantially outperforms image-space positional encodings, which are sensitive to the camera pose.
Together, these give a policy the geometric structure that makes 3D approaches robust while retaining the scalability and pretrained priors of a 2D image backbone.
We refer to the combination of these ingredients as \method, a \textbf{\underline{V}}iewpoint-\textbf{\underline{G}}eneralizable manipulation \textbf{\underline{P}}olicy that remains performant across views in the training camera distribution. \method outperforms 2D and 3D alternatives under randomized camera placement, while approaching fixed-camera performance.  %
\method is deliberately not a new architecture. Each of its components appears in prior work; our contribution is to isolate the pieces responsible for viewpoint robustness, under a common dataset and action head. In summary, we make the following contributions:
 
\begin{enumerate}[label=(\arabic*),leftmargin=*]
    \item We present a controlled empirical study of which design choices allow visuomotor policies to generalize across viewpoints.
    \item We show that grounding all tokens (vision, proprioception, action) in the robot frame via rotary positional encodings substantially outperforms image-space positional encodings.
    \item We demonstrate the practical utility of this recipe by training on diverse viewpoints in simulation and transferring the resulting policy zero-shot to the real world under random cameras.
\end{enumerate}

\section{Related Work}

\textbf{Viewpoint Generalizable Manipulation.} A growing body of work studies how to effectively train viewpoint generalizable manipulation policies~\cite{jiang2026knowyourcamera, tian2024vista, kleeraven, yuan2024learning, Wang2025articubot}. One line of work tackles the problem through \emph{data}, such as by collecting demonstrations spanning heterogeneous views~\cite{khazatsky2024droid}, rendering random views in simulation~\cite{Wang2025articubot, deshpande2026molmobot}, or augmenting fixed-camera data with synthesized novel-view observations~\cite{tian2024vista}. A second family targets the \emph{representation}, learning view-robust features through self-supervision~\cite{seo2023multi} or contrastive objectives~\cite{lee2025class, yuan2024learning}. Yet another line of work is \emph{camera pose conditioning}, which feeds extrinsics directly to the policy. Jiang \etal\cite{jiang2026knowyourcamera} directly encode camera poses as Pl\"ucker raymaps as additional policy input.
While camera conditioning provides information on where the camera is, our intuition is that the policy needs to know the location of the \textit{observed pixels} with respect to the \textit{robot}, not the location of the camera itself.
Our work also relates to recent progress in transferring RGB-based manipulation policies from simulation to the real world~\cite{yin2026emergent, singh2025dextrah, ankile2025imitation, deshpande2026molmobot, xue2025opening, yuan2024learning}. Many methods for visual sim-to-real assume known camera poses prior to data generation in simulation~\cite{yin2026emergent, xue2025opening}, or apply randomization around a narrower range of camera poses~\cite{ankile2025imitation, yuan2024learning}. More closely related to our work, Yuan \etal\cite{yuan2024learning} and Deshpande \etal\cite{deshpande2026molmobot} both study viewpoint generalization in the context of sim-to-real transfer. Deshpande \etal approach
the problem by training a massive VLA at scale. In contrast, we ask which architectural choices promote viewpoint robustness, and show that a policy built from those choices transfers zero-shot to the real world under random camera configurations.
 
\textbf{Geometry-Aware Representations for Manipulation.} A complementary line of work adds 3D structure to the policy, but methods differ in \emph{where} that structure enters. Policies over point clouds or voxels~\cite{Ze2024DP3, shridhar2023perceiver} are viewpoint-robust by construction, since their inputs can be transformed to a scene-centric frame, but they forgo image-scale pretraining and typically downsample or voxelize the observations to remain tractable, thereby losing fine image details. Reprojection methods~\cite{goyal2023rvt, goyal2024rvt, yang2026remap} instead project multi-view RGB-D onto pre-defined canonical views, but the holes that the projection introduces cause the input to be out-of-distribution for pretrained image backbones. Another route keeps the image pipeline and uses a 3D foundation model for \emph{features}: VGGT-DP~\cite{ge2025vggt} pairs a VGGT encoder~\cite{wang2025vggt} with a diffusion policy but degrades sharply under modest camera rotations, indicating that geometry-aware features alone are not sufficient for viewpoint robustness. Closest to us, 3D Diffuser Actor (3DDA)~\cite{ke20243d} and 3D FlowMatch Actor (3DFA)~\cite{gkanatsios20253d} lift features and actions into a shared 3D frame and reason over them jointly.
Our work shares the approach of grounding action generation in 3D, but focuses on the challenge of robustness to large changes in camera viewpoint. While 3DFA combines several of the choices we study, its experiments do not isolate their effects on viewpoint robustness. Our contribution is to identify the key design choices that contribute to viewpoint generalization.

\section{Problem Statement}
 
\textbf{Visual Imitation Learning.} We consider the standard visual imitation learning setting, in which a policy $\pi_\theta$ is trained to imitate expert behavior from a dataset of demonstrations $\mathcal{D} = \{d_1, \dots, d_N\}$. Each trajectory is a sequence of timestep tuples $d_i = ( (o_1, s_1, a_1), \dots, (o_{T_i}, s_{T_i}, a_{T_i}) )$, where $s_t \in \mathcal{S}$ denotes the proprioceptive robot state (\ie end-effector pose and gripper width), $a_t \in \mathcal{A}$ denotes the expert action, where $\mathcal{A}$ is a delta end-effector action space (translation, rotation, and gripper command), and $o_t$ denotes the visual observation at timestep $t$. A visual observation consists of one or more calibrated RGB-D views, $o_t = \{ ( I_t^{(v)}, D_t^{(v)}, K_t^{(v)}, E_t^{(v)} ) \}_{v=1}^{V}$, where $V$ is the number of cameras and, for each view $v$, $I_t^{(v)} \in \mathbb{R}^{H \times W \times 3}$ is an RGB image, $D_t^{(v)} \in \mathbb{R}^{H \times W}$ is the corresponding metric depth map, $K_t^{(v)} \in \mathbb{R}^{3 \times 3}$ is the camera intrinsics matrix, and $E_t^{(v)} \in SE(3)$ is the camera extrinsics expressed in the robot's base frame.
 
\textbf{Viewpoint Generalizable Policy Learning.} We assume that the camera pose of each trajectory $d_i$ during training is independently drawn from a distribution over camera poses, so that the demonstration dataset $\mathcal{D}$ spans a wide range of viewpoints. At deployment, the camera pose is likewise sampled from the same camera pose distribution. We aim to learn a single policy whose performance is consistent across the full distribution of camera poses.
 
\textbf{Assumptions.} We assume that all cameras are calibrated, such that intrinsics $K_t^{(v)}$ and extrinsics $E_t^{(v)}$ are available during both training and deployment. Second, we assume that each RGB image is paired with a depth map $D_t^{(v)}$, supplied either by a depth sensor or by an RGB-based depth estimator.
 
\begin{figure*}[!t]
  \centering
  \includegraphics[width=\textwidth]{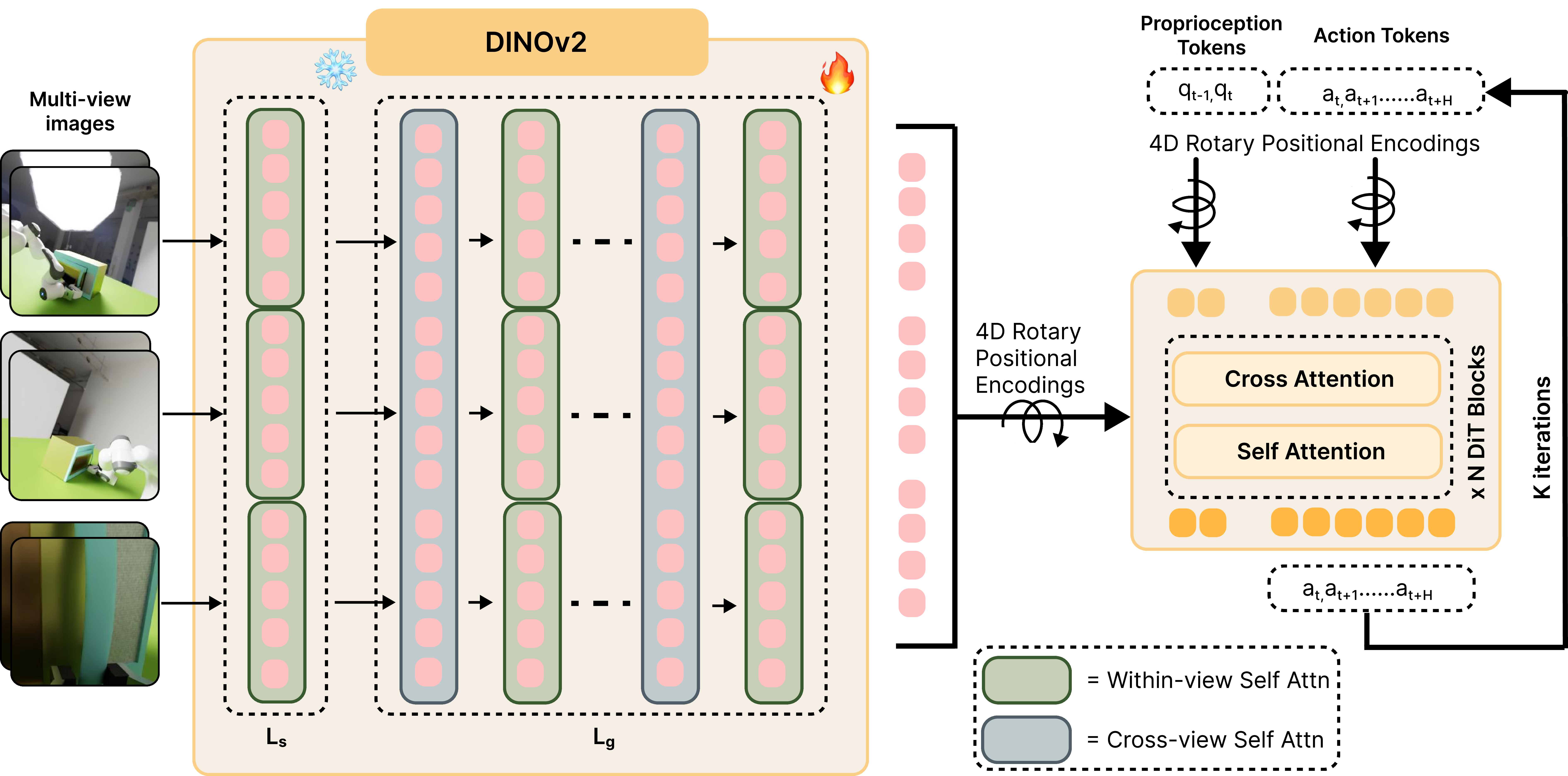}
  \caption{\textbf{Base policy and design axes}, shown in the \method configuration. A set of calibrated RGB views are encoded by a backbone initialized from DINOv2 applying within-view self-attention and cross-view global self-attention, producing dense multiview image tokens (\emph{visual encoder} and \emph{token compression} axes). Every image, proprioception, and action token is assigned a spatiotemporal position $(x, y, z, h)$ and attended over with a 4D rotary positional encoding (\emph{positional encoding} axis). An $N$-block DiT, shared across all variants, cross-attends from proprioception/action tokens to image tokens and self-attends over the action chunk.}
  \label{fig:method}
\end{figure*}

\section{Study Design}\label{sec:study_design}

To identify what matters for viewpoint generalization, we fix a base policy, training data, and evaluation protocol, and vary one design choice at a time. We organize these choices along three axes, each targeting how a policy should process observations from changing viewpoints: (i)~whether visual features should be compressed before the action head, (ii)~which properties of the visual encoder matter, and (iii)~how token positions should be encoded. Fig.~\ref{fig:method} illustrates the base policy in the \method configuration.

\subsection{Base Policy}\label{sec:base_policy}
We first describe the base policy that provides the common foundation for evaluating each design choice.
All variants follow a standard design for diffusion or flow-matching based policies for the action head, using a Diffusion Transformer (DiT)~\cite{peebles2023scalable} and interleaving cross-attention to observation tokens with self-attention over the action sequence~\cite{alayrac2022flamingo, jiang2023vima, nvidia2025groot, black2024pi_0}. At each prediction step, the policy takes a history of $T_o$ observation steps as input and predicts an action sequence with $T_a$ steps. An MLP encodes the proprioceptive state $s_t$ and the noised actions into tokens. Each block cross-attends from these tokens to the image tokens, then self-attends among proprioception and action tokens.

\textbf{Action generation.} Following~\cite{nvidia2025groot, black2024pi_0}, we generate action chunks with a flow matching training objective. Starting from a ground-truth action chunk $a$ and Gaussian noise $\epsilon \sim \mathcal{N}(0, I)$, we define the interpolated action chunk $a^\tau = (1-\tau)\,\epsilon + \tau\,a$ where $\tau \in \mathcal{U}[0, 1]$. We train the action head to predict the velocity field $\frac{da_\tau}{d\tau} = a - \epsilon$ by minimizing the loss $\mathcal{L} = \lVert v_\theta(a_\tau, \tau) - (a - \epsilon) \rVert^2$. At inference, we integrate the predicted velocity from noise to a clean action chunk with Euler integration.

\subsection{Design Axes}\label{sec:design_axes}

\textbf{Axis 1: Visual token compression.} Some policies compress visual observations into a single global feature vector before the action head~\cite{chi2023diffusionpolicy, Ze2024DP3}, while others retain a set of visual tokens that the action head attends to via cross-attention~\cite{nvidia2025groot, zhao2023learning, ke20243d}. We compare retaining all dense patch tokens against global compression via spatial softmax or max pooling.

\textbf{Axis 2: Visual encoder.} We compare different visual encoders: a ResNet-18, a frozen DINOv2~\cite{oquab2023dinov2}, a finetuned DINOv2, and a DINOv2 finetuned with cross-view attention. %
The cross-view variant follows the architecture of Depth Anything 3~\cite{lin2025depth}, building on recent advances in multi-view visual estimation~\cite{wang2025vggt, lin2025depth}. The vision transformer is divided such that the first $L_s$ layers apply within-view self-attention, while the subsequent $L_g$ layers alternate between within-view self-attention and cross-view global self-attention. Following Lin \etal\cite{lin2025depth}, we use $L_s + L_g = 12$ transformer blocks and alternate between the two attention mechanisms for the final $L_g = 8$ blocks. Rather than fully finetuning DINOv2, we freeze the first $L_s$ within-view layers and finetune the remaining blocks, which reduces computational cost while initializing from pretrained features.

\textbf{Axis 3: Positional encoding.} Transformer-based policies typically encode token positions with learned or sinusoidal encodings over image-space patch indices~\cite{zhao2023learning, nvidia2025groot}, which carry no consistent geometric meaning when the camera moves. We compare image-space sinusoidal encodings against rotary positional encodings (RoPE)~\cite{su2021roformer} computed from each token's location in the robot base frame, using its 3D position and timestep (4D RoPE). Approaches that instead condition the policy on camera geometry, Pl\"ucker raymaps~\cite{jiang2026knowyourcamera} and canonical views~\cite{goyal2023rvt, yang2026remap}, are evaluated as alternatives in Sec.~\ref{sec:main_results}.

\textbf{Robot-frame token positions.} Each image, proprioception, and action token is assigned a 4D coordinate $(x,y,z,h)$, where $x,y,z$ denote the position in the robot base frame and $h$ denotes the token's temporal index in the history window relative to the current timestep. We implement 4D RoPE by partitioning each attention head into separate $x$, $y$, $z$, and $h$ subspaces and applying standard 1D rotary positional embeddings independently to each coordinate dimension. For observation tokens, $h \in \{0,\dots,T_o-1\}$ is the observation-history index within the $T_o$-frame history window, and likewise for proprioception tokens. Image-token 3D positions are obtained by lifting pixels using the depth map and camera pose. We take the 3D coordinate from the center pixel of each patch as the token's position~\cite{krishna2026ghost}. Proprioception-token positions are given by the gripper's 3D position.

Action tokens have no ground-truth position at inference. Similar to~\cite{ke20243d, gkanatsios20253d}, we represent actions as relative gripper displacements, and estimate the future gripper position at action timestep $t_a$ by integrating the translational action components up to that timestep: $\hat{g}_{t_a}^{\tau} = g + \sum_{j=0}^{t_a} \Delta_j(a^\tau)$,
where $g$ is the current gripper position, $a^\tau$ is the noised action chunk at flow time $\tau$, and $\Delta_j(a^\tau)$ is the translational displacement component of the $j$-th action. Early in denoising ($\tau\approx 0$) this estimate is dominated by noise and the action tokens cluster near arbitrary positions; as the chunk denoises it converges to the realized future gripper trajectory, so action tokens are progressively grounded at the locations they will act on.

\subsection{\texorpdfstring{\method}{VGP} and Reference Methods}\label{sec:full_recipe}

We refer to the combination of dense tokens, a cross-view finetuned DINOv2 encoder, and robot-frame 4D RoPE as \method. Each ablation in Sec.~\ref{sec:experiments} changes a single axis relative to \method while keeping all other components fixed. We additionally report Diffusion Policy~\cite{chi2023diffusionpolicy}, ACT~\cite{zhao2023learning}, and DP3~\cite{Ze2024DP3} as external reference points; these differ from the base policy in more than one component.

\begin{figure}[!t]
  \centering
  \includegraphics[width=\columnwidth]{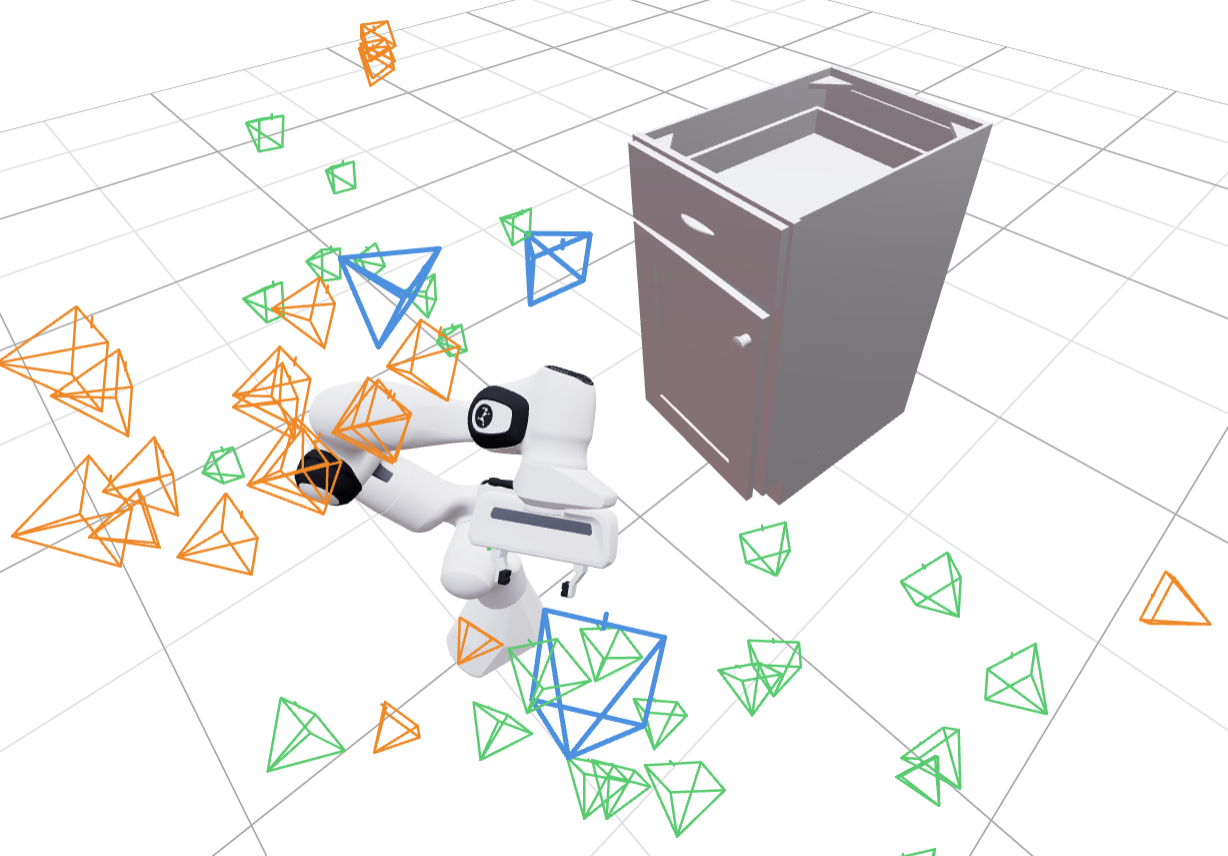}
  \caption{\textbf{Camera Distribution.} We visualize the camera distribution (\textcolor{figgreen}{\textbf{green}}), the poses of the cameras used in the real world (\textcolor{figblue}{\textbf{blue}}) (Sec. \ref{sec:sim2real}), and the poses used for evaluating out-of-distribution generalization (\textcolor{figorange}{\textbf{orange}}) (Sec. \ref{sec:ood_cam_pose}).}
  \label{fig:cam_pose}
\end{figure}

\section{Experiments and Analysis}\label{sec:experiments}
 
We evaluate \method on simulated tasks under randomized cameras for a controlled comparison against ablations of each design choice. We then demonstrate zero-shot sim-to-real transfer on a physical robot with camera poses that are unknown during simulation data generation.
 
\textbf{Experimental Setup.} We compare \method against viewpoint generalization methods in tasks proposed by Jiang \etal\cite{jiang2026knowyourcamera}, as well as a custom articulated object manipulation task in PyBullet, inspired by Wang \etal\cite{Wang2025articubot}. In contrast to Jiang \etal\cite{jiang2026knowyourcamera}, we include proprioception in the policy's input and for each method we report the success rate of its best checkpoint, rather than averaging the final ten checkpoints. We report results averaged across three seeds.  
 
For the articulated object opening task, we train a policy to imitate an expert motion planner opening a cabinet with a small knob handle from the PartNet-Mobility dataset~\cite{xiang2020sapien}. In contrast to front-facing viewpoint setups that look towards the robot~\cite{ankile2025imitation, yuan2024learning, jiang2026knowyourcamera}, we randomly sample two over-the-shoulder camera views, from which the robot is harder to localize from visual observation alone (see Fig.~\ref{fig:cam_pose}). We include a wrist-mounted camera, which provides a close-up view of the manipulated object and ensures the object remains visible even when the robot occludes it from the over-the-shoulder views. We generate 989 demonstration trajectories with over 100k observation-action pairs. For evaluation, we separately generate 100 test trajectories and sample random camera poses for each test case. As in prior work~\cite{Wang2025articubot}, we use the normalized opening performance (``Norm.\ Open'') as our evaluation metric, which measures the ratio of opened joint angle between the method and the expert demonstrator. 

\subsection{Camera information}\label{sec:main_results}
 
\begin{table*}[!t]
\centering
\caption{We compare various approaches to incorporating camera information for viewpoint generalization on opening an articulated cabinet and on the ``Do You Know Where Your Camera Is?'' benchmark. Unless marked as fixed camera, cameras are randomized during training and evaluation. Mean $\pm$ std.\ over 3 seeds. Avg.\ is computed over Cabinet, Pick Place Can, and Square Assembly. Best: \colorbox{gray!25}{\textbf{shaded}}, second-best: \underline{underline}. $^*$Under the fixed camera setting, the canonical views coincide with the fixed camera views, so Canonical Views receives the same inputs and uses the same architecture as the ``w/ Sinusoidal'' ablation (Table~\ref{tab:ablations}); the two are identical, and we report the same result.}
\label{tab:main_results}
\footnotesize
\setlength{\tabcolsep}{4pt}
\begin{tabular}{l c|c cc c}
\toprule
 & \multicolumn{1}{c|}{Fixed camera} & \multicolumn{4}{c}{Randomized cameras} \\
\cmidrule(lr){2-2} \cmidrule(lr){3-6}
 & \multicolumn{2}{c}{\begin{tabular}{@{}c@{}}Articulated Objects\\(Norm.\ Open)\end{tabular}}
 & \multicolumn{2}{c}{\begin{tabular}{@{}c@{}}Do You Know Where Your Camera Is?\\(Success \%)\end{tabular}} & \\
\cmidrule(lr){2-3} \cmidrule(lr){4-5}
Method & Cabinet & Cabinet & Pick Place Can & Square Assembly & Avg. \\
\midrule
Pl\"ucker Raymaps & 63.6 \std{2.9} & 39.5 \std{1.4} & \underline{68.0 \std{0.0}} & 22.7 \std{1.2} & 43.4 \\
Canonical Views & \underline{59.1 \std{3.2}}$^*$ & \underline{52.8 \std{3.2}} & 62.0 \std{0.0} & \colorbox{gray!25}{\textbf{32.0 \std{0.0}}} & \underline{48.9} \\
4D RoPE (\method) & \colorbox{gray!25}{\textbf{69.1 \std{0.8}}} & \colorbox{gray!25}{\textbf{64.4 \std{0.8}}} & \colorbox{gray!25}{\textbf{96.7 \std{3.1}}} & \underline{29.3 \std{3.1}} & \colorbox{gray!25}{\textbf{63.5}} \\
\bottomrule
\end{tabular}
\end{table*}

First, we compare various approaches for incorporating camera information into the policy.
To ensure a controlled comparison, we keep as much of the architecture and training procedure as similar as possible.  All  approaches that we evaluate use the same visual encoder and the same action head architecture.  For positional encoding,  \method uses 4D RoPE (as described in Sec.~\ref{sec:design_axes}), whereas the other approaches use 
additive sinusoidal positional embeddings. 
 
\textbf{Pl\"ucker-Conditioned Policies}~\cite{jiang2026knowyourcamera} condition the imitation learning policy on Pl\"ucker raymaps provided as auxiliary inputs. We train a vision transformer from scratch to encode the Pl\"ucker raymaps. We adopt the \textit{late fusion} strategy, which Jiang \etal \cite{jiang2026knowyourcamera} identify as the most effective way to integrate the two streams. Each Pl\"ucker token is concatenated channel-wise with its corresponding DINOv2 image token and projected back to the token dimension. The resulting fused tokens condition the DiT action head via cross-attention.
 
\textbf{Canonical Views}~\cite{goyal2023rvt, goyal2024rvt, yang2026remap} canonicalize the input by lifting RGB images into 3D using depth and camera extrinsics, then reproject the resulting colored point cloud into \textit{virtual images} rendered from a fixed set of camera poses, using z-buffering to resolve occlusions. For the Cabinet task, we fix the virtual cameras to the left and right of the Franka arm to imitate over-the-shoulder views. For the tasks from Jiang \etal\cite{jiang2026knowyourcamera}, we select two arbitrary views from the dataset of training camera poses as canonical viewpoints.

\textbf{Results.} Table~\ref{tab:main_results} reports our main comparison. As shown, using 4D RoPE outperforms or performs competitively with the alternatives across all tasks. 
 
\subsection{Design choice analysis}
 
We perform further ablations to isolate the contribution of each design choice. All ablations are conducted on the Cabinet, Pick Place Can, and Square Assembly tasks with randomized camera placement during both training and evaluation, unless mentioned otherwise.
In the following sections, we present our findings and provide insights into what factors matter for viewpoint generalizable policy learning.
 
\textbf{Axis 1: Visual token compression.} To isolate the effect of compression within our architecture, we ablate \method by replacing its token set with global compression (spatial softmax or max pooling) before the action head. Additionally, we report the performance of Diffusion Policy as an external baseline, which uses spatial softmax (but with a different architecture). Table~\ref{tab:ablations} shows that global compression performs reasonably under fixed cameras but causes a large drop in performance under randomized cameras, where the action head loses the spatial information needed to reason across viewpoints. 

\begin{table*}[!t]
\centering
\caption{Ablations of each design choice, varied relative to \method: visual token compression, visual encoder, and positional encoding. The first task column uses a fixed camera; the remaining task columns use cameras randomized during training and evaluation. Mean $\pm$ std.\ over 3 seeds. Avg.\ is computed over the three randomized-camera tasks: Cabinet, Pick Place Can, and Square Assembly. \method retains all image tokens, uses a cross-view finetuned DINOv2 encoder, and uses robot-frame 4D RoPE; rows marked ``w/'' replace a single component of \method. Best: \colorbox{gray!25}{\textbf{shaded}}, second-best: \underline{underline}.}
\label{tab:ablations}
\footnotesize
\setlength{\tabcolsep}{4pt}
\begin{tabular}{l c|c cc c}
\toprule
 & \multicolumn{1}{c|}{Fixed camera} & \multicolumn{4}{c}{Randomized cameras} \\
\cmidrule(lr){2-2} \cmidrule(lr){3-6}
 & \multicolumn{2}{c}{\begin{tabular}{@{}c@{}}Articulated Objects\\(Norm.\ Open)\end{tabular}}
 & \multicolumn{2}{c}{\begin{tabular}{@{}c@{}}Do You Know Where Your Camera Is?\\(Success \%)\end{tabular}} & \\
\cmidrule(lr){2-3} \cmidrule(lr){4-5}
Method & Cabinet & Cabinet & Pick Place Can & Square Assembly & Avg. \\
\midrule
\multicolumn{6}{l}{\underline{\textit{Token compression:}}}\\
Diffusion Policy          & 56.1 \std{3.2} & 33.2 \std{4.3} & 1.3 \std{1.2} & 2.7 \std{1.2} & 12.4 \\
w/ Spatial Softmax        & 55.2 \std{8.4} & 33.4 \std{0.2} & 56.0 \std{5.3} & 22.7 \std{1.2} & 37.4 \\
w/ Max Pooling            & 60.0 \std{3.9} & 35.1 \std{1.9} & 61.3 \std{4.2} & 22.7 \std{3.1} & 39.7 \\
\midrule
\multicolumn{6}{l}{\underline{\textit{Visual encoder:}}}\\
w/ ResNet-18              & 67.2 \std{0.6} & \underline{60.9 \std{2.7}} & 84.7 \std{4.6} & 26.0 \std{5.3} & 57.2 \\
w/ DINOv2 (frozen)        & 63.6 \std{1.4} & 59.8 \std{0.6} & 88.0 \std{2.0} & 29.0 \std{9.6} & 58.9 \\
w/ DINOv2 (finetuned)     & \underline{67.6 \std{0.9}} & 59.3 \std{1.5} & \underline{93.3 \std{5.8}} & \colorbox{gray!25}{\textbf{32.0 \std{2.0}}} & \underline{61.5} \\
\midrule
\multicolumn{6}{l}{\underline{\textit{Positional encoding:}}}\\
ACT                       & 58.9 \std{3.1} & 39.2 \std{1.8} & 35.3 \std{1.2} & 20.0 \std{2.0} & 31.5 \\
w/ Sinusoidal             & 59.1 \std{3.2} & 39.5 \std{3.5} & 68.7 \std{4.2} & 24.0 \std{3.5} & 44.1 \\
\midrule
\method               & \colorbox{gray!25}{\textbf{69.1 \std{0.8}}} & \colorbox{gray!25}{\textbf{64.4 \std{0.8}}} & \colorbox{gray!25}{\textbf{96.7 \std{3.1}}} & \underline{29.3 \std{3.1}} & \colorbox{gray!25}{\textbf{63.5}} \\
\bottomrule
\end{tabular}
\end{table*}

\begin{figure*}[!t]
\centering
\begin{minipage}[t]{0.48\textwidth}
\centering
\includegraphics[width=\linewidth]{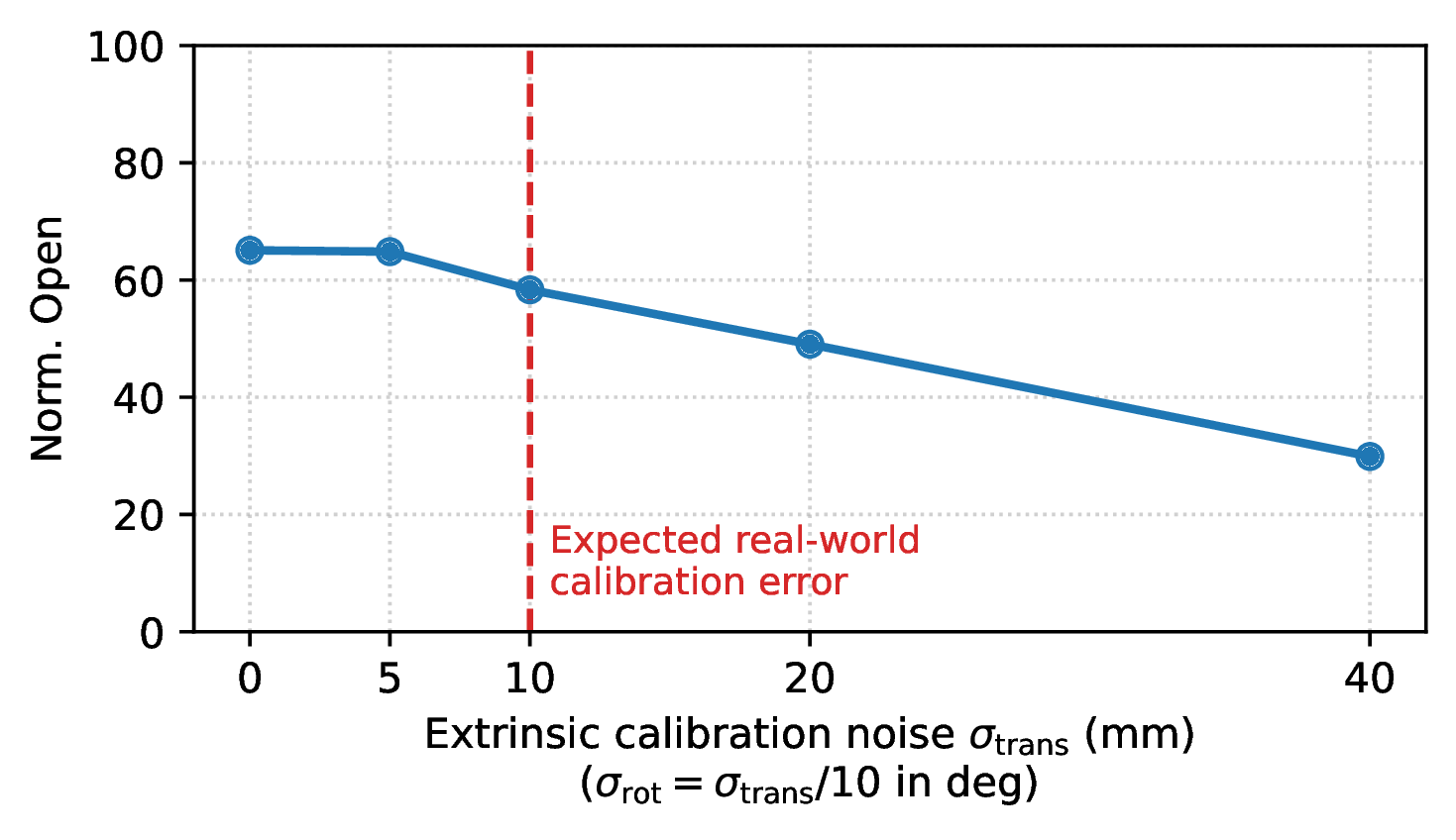}
\caption{Cabinet (Norm.\ Open) under extrinsic calibration noise at evaluation. The dashed line marks the calibration error we expect in the real world ($\sigma_{\text{trans}} \approx 1$\,cm, $\sigma_{\text{rot}} \approx 1^\circ$). The policy is trained without calibration noise.}
\label{fig:noise_sweep}
\end{minipage}\hfill
\begin{minipage}[t]{0.48\textwidth}
\centering
\includegraphics[width=\linewidth]{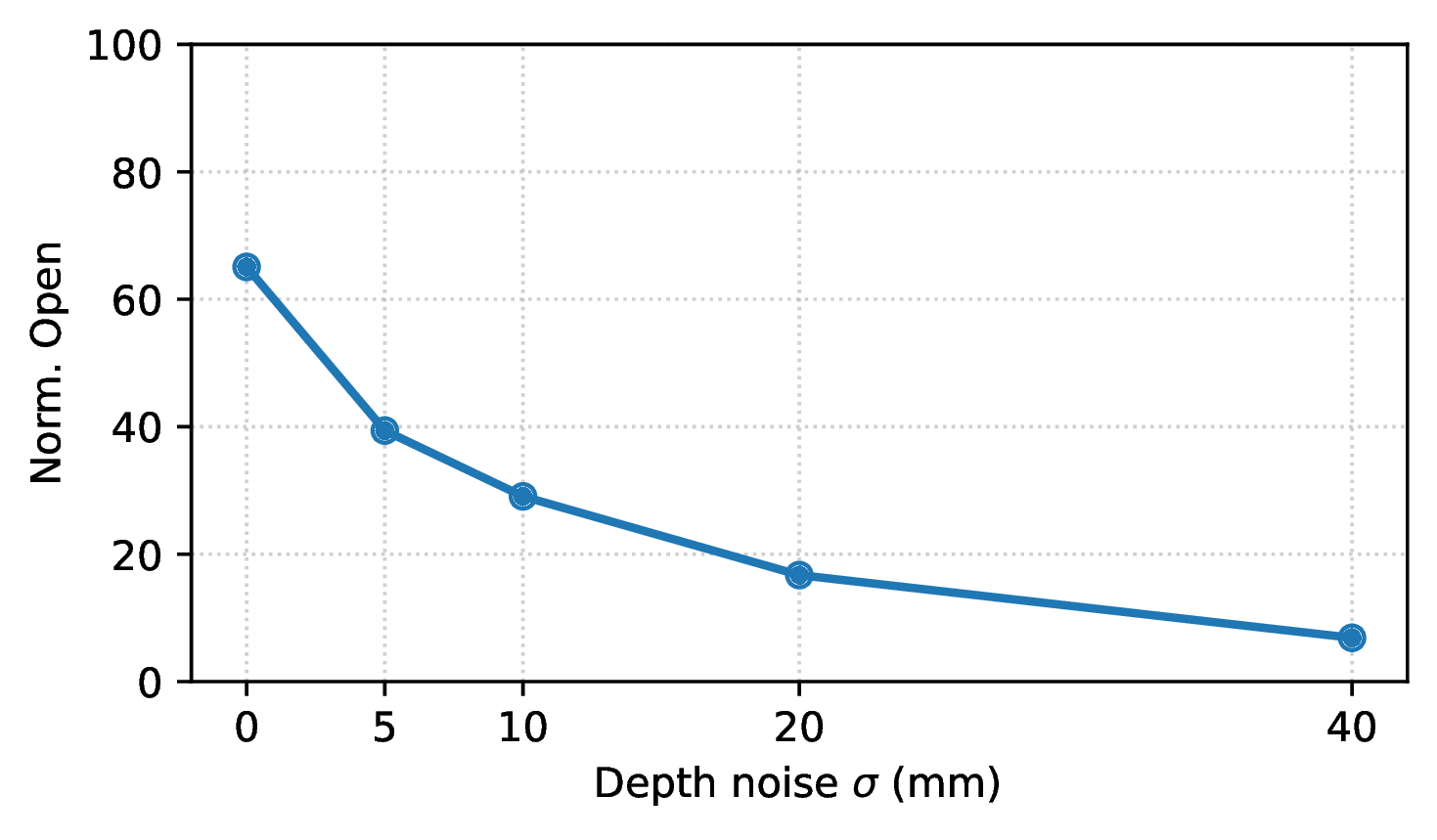}
\caption{Cabinet (Norm.\ Open) under additive Gaussian depth noise at evaluation. The policy is trained without depth noise.}
\label{fig:depth_noise}
\end{minipage}
\end{figure*}

\begin{figure}[!t]
\centering
\includegraphics[width=\columnwidth]{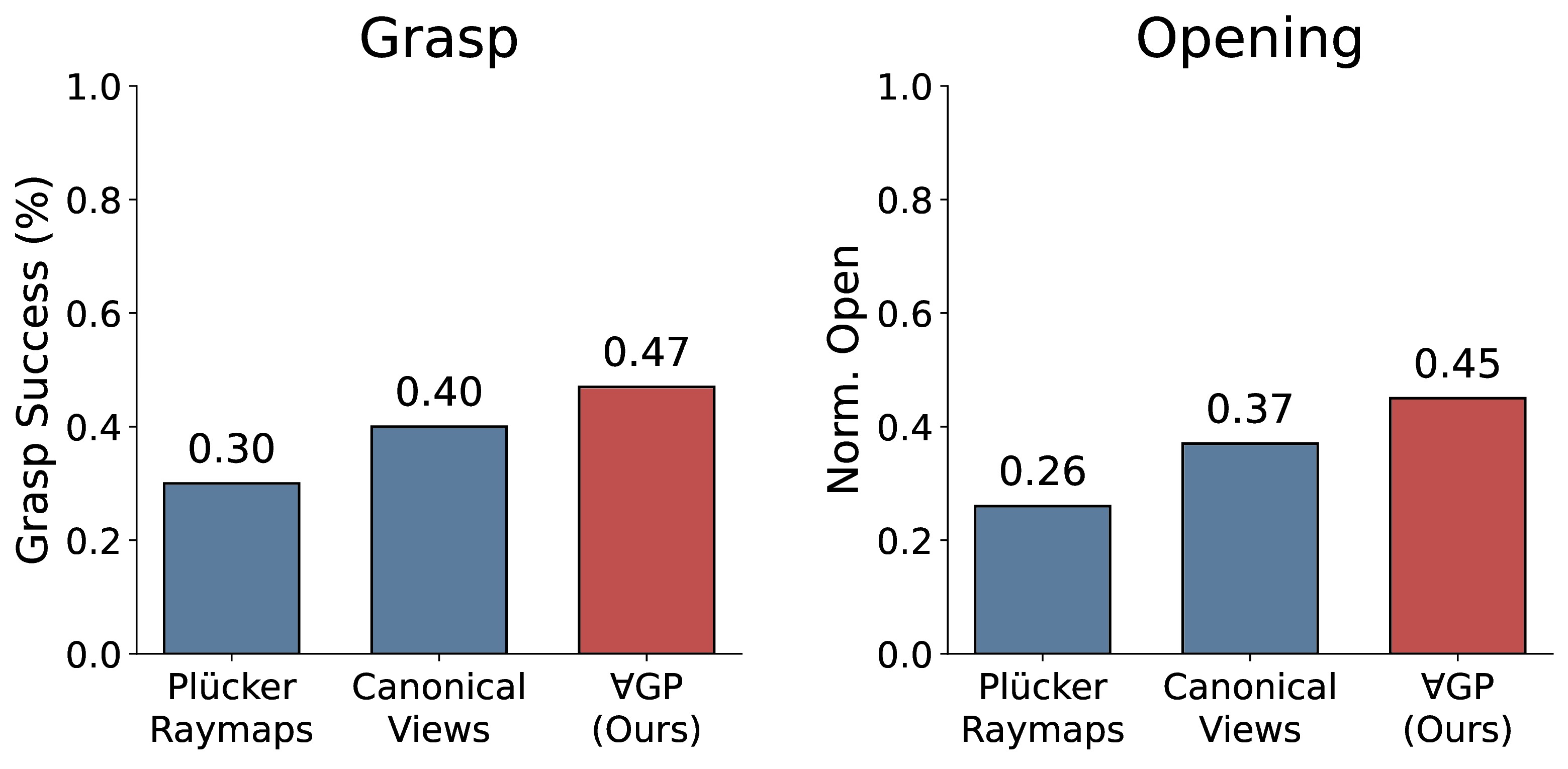}
\caption{Sim-to-real transfer on microwaves. Policies trained in simulation and evaluated zero-shot in real world under randomly sampled camera poses. Performance averaged over 30 trials across 3 camera configurations (10 trials per camera configuration).}
\label{fig:sim2real_microwaves}
\end{figure}

\textbf{Axis 2: Visual encoder.}\label{sec:abl_encoder} To understand the effects of visual encoders on viewpoint generalization, we evaluate \method with several modifications of the visual encoder. Our results are shown in Table~\ref{tab:ablations}. %
Our results indicate that cross-view attention layers, as employed in prior work for multi-view depth estimation~\cite{lin2025depth, wang2025vggt}, yield additional gains %
by helping the encoder fuse information across cameras. However, we find that the performance differences from visual encoder changes are less significant than the gains from removing token compression or changing the positional encoding.
 
\textbf{Axis 3: Positional encoding.} \label{sec:abl_rope} Transformer-based policies typically inject positional information for image tokens via learned or sinusoidal encodings over patch (row, column) indices~\cite{zhao2023learning, nvidia2025groot, vaswani2017attention}, leaving the action head to recover scene geometry implicitly from pixel content. We find that this is fragile under randomized cameras because the same patch index can correspond to different 3D locations as the camera moves, so the encoding carries no consistent geometric meaning. Table~\ref{tab:ablations} shows that replacing image-space encodings with geometrically grounded, robot-frame positional encodings via 4D RoPE improves performance substantially from $39.5$ to $64.4$ normalized opening on Cabinet with randomized cameras. The geometric relationship between any visual token and the gripper is, under this encoding, invariant to the camera pose.

\subsection{Sim-to-Real Transfer}\label{sec:sim2real}
 
We validate the practical usefulness of \method on a sim-to-real transfer task involving articulated object manipulation. We build on Wang \etal\cite{Wang2025articubot}, which provides an automated pipeline for generating demonstrations of a Franka robot opening articulated objects in simulation. From the 322 objects used in~\cite{Wang2025articubot}, we select five microwaves, generate a total of 2358 demonstrations and over 100{,}000 state-action pairs, and render them in IsaacLab~\cite{mittal2025isaaclab}.
 
\textbf{Domain Randomization.}
To bridge the sim-to-real gap, we apply extensive domain randomization to the rendered outputs. Following prior work~\cite{yin2026emergent, singh2025dextrah}, we randomize all aspects of the visual scene, including microwave colors and textures, the robot, the table, and the background. We additionally sample camera poses per trajectory and render each trajectory from two over-the-shoulder views and one wrist camera view, with camera intrinsics matched to the ZED2 and ZED-mini cameras used in our real-world experiments. Depth maps are augmented to simulate noise from depth sensors, including edge artifacts and sensor noise. All approaches are trained on the same domain-randomized dataset to ensure that performance gaps are due to architectural changes only. 

\textbf{Real-World Deployment.} We deploy trained policies on a Franka robot arm. The camera poses are not known during data generation but are within distribution of the randomized camera poses used in simulation. The training distribution and the real-world camera poses are visualized in Fig.~\ref{fig:cam_pose}. 
Real-world depth is obtained using Fast-FoundationStereo~\cite{wen2026fastfoundationstereo}, and depth beyond the robot workstation is cropped out using a bounding box. We apply this crop to match the training data distribution since ground-truth background depth is not available during training. We mount three cameras and conduct 10 trials for each of the three possible camera pairs (30 trials per policy) (see Fig.~\ref{fig:teaser}). We report both the grasp success rate, visually confirmed in every trial, and the normalized opening performance, defined as the ratio between the object's opening distance and the maximal opening distance of the object subject to robot joint limits and workspace constraints~\cite{Wang2025articubot}. As shown in Fig.~\ref{fig:sim2real_microwaves}, \method outperforms Canonical Views and Pl\"ucker Raymaps on both grasp success rate and opening performance, validating that the findings from simulation carry over to the real world. %

\subsection{Generalization to out-of-distribution viewpoints}\label{sec:ood_cam_pose}

In the above analysis, the test camera poses are sampled from the same distribution as the training set to evaluate viewpoint \emph{robustness}. %
In this section, we evaluate with an even broader distribution of camera poses to evaluate the viewpoint \textit{generalization} abilities of the policies. In Fig.~\ref{fig:cam_pose} we visualize camera poses from this broader distribution, which observe the cabinet from unseen elevated and oblique angles. 
Our results are shown in Table~\ref{tab:ood_poses}. While all policies drop in performance compared to the \textit{in-distribution} Randomized Camera results from Table~\ref{tab:main_results}, \method retains its advantage over other camera-conditioned approaches under unseen viewpoints. 

\begin{table}[!t]
\centering
\caption{Evaluation on the \textbf{Cabinet} task with \textbf{out-of-distribution cameras}. Best: \colorbox{gray!25}{\textbf{shaded}}, second-best: \underline{underline}.}
\label{tab:ood_poses}
\footnotesize
\setlength{\tabcolsep}{4pt}
\begin{tabular}{l c}
\toprule
Method & \textbf{Cabinet} (Norm.\ Open) \\
\midrule
Pl\"ucker Raymaps & 29.8 \std{5.2} \\
Canonical Views & \underline{45.6 \std{2.3}} \\
\method & \colorbox{gray!25}{\textbf{51.4} \std{5.2}} \\
\bottomrule
\end{tabular}
\end{table}

\subsection{Comparison to 3D Policies}
 
The methods evaluated above all operate on RGB or lifted-RGB features. Here we additionally compare against a \emph{pure} 3D policy that consumes a point cloud directly. %
We use 3D Diffusion Policy (DP3)~\cite{Ze2024DP3}, %
following the DP3 implementation provided by Wang \etal\cite{Wang2025articubot}. For DP3, at each timestep, we unproject the depth from all calibrated cameras, segment the object using the ground-truth segmentation mask, and fuse the resulting points into a single object point cloud in the robot base frame. To match the observation space of \method, the object point cloud is fused from the two randomly sampled views and a wrist view. The point cloud is downsampled to $4{,}500$ points via farthest-point sampling. As in Wang \etal\cite{Wang2025articubot}, we incorporate a $4$-point gripper point cloud as the proprioceptive state as input to DP3. As shown in Table~\ref{tab:dp3_comparison}, DP3 exhibits a small gap between fixed and randomized cameras, confirming that a point-cloud based policy is viewpoint-robust by construction. \method achieves a comparably small gap and exceeds DP3 in performance.
 
\begin{table}[!t]
\centering
\caption{Comparison against DP3 on the \textbf{Cabinet} task. Best: \colorbox{gray!25}{\textbf{shaded}}.}
\label{tab:dp3_comparison}
\footnotesize
\setlength{\tabcolsep}{4pt}
\begin{tabular}{l cc}
\toprule
Method & \multicolumn{2}{c}{\textbf{Cabinet} (Norm.\ Open)} \\
\cmidrule(lr){2-3}
       & Fixed camera & Randomized cameras \\
\midrule
DP3~\cite{Ze2024DP3} (point cloud, GT seg.) & 50.9 \std{2.0} & 43.9 \std{1.4} \\
\method  & \colorbox{gray!25}{\textbf{69.1} \std{0.8}} & \colorbox{gray!25}{\textbf{64.4} \std{0.8}} \\
\bottomrule
\end{tabular}
\end{table}

\subsection{Robustness to Calibration and Depth Noise}
 
\method lifts pixels into the robot frame using camera extrinsics and depth, both of which are imperfect in the real world. We characterize its sensitivity to such noise by evaluating a trained policy on the \emph{Cabinet} task under increasing test-time perturbation of the extrinsics and depth. For the extrinsics, we add zero-mean Gaussian noise to the translational and rotational components (Fig.~\ref{fig:noise_sweep}). For depth, we add zero-mean Gaussian noise to the input depth maps (Fig.~\ref{fig:depth_noise}). Importantly, the policy is trained without depth or calibration error. As shown in Fig.~\ref{fig:noise_sweep} and Fig.~\ref{fig:depth_noise}, \method exhibits robustness against reasonable calibration error, but tends to degrade in performance as depth noise increases. Training the model with extrinsic or depth noise would likely improve robustness.

\section{Conclusion}

In this work, we investigate the problem of \textit{viewpoint generalization}, relaxing the common constraint of fixed cameras at training and deployment time. Our experiments identify key factors that drive viewpoint generalization: retaining dense scene representations and grounding every visual, proprioception, and action token in the geometry of the scene. We instantiate our recommendations as \method, a policy that is performant across variations in camera viewpoints. 
Finally, we demonstrate a practical use-case of these recommendations by showcasing zero-shot sim-to-real transfer under random cameras. We hope this study offers useful insights for developing viewpoint-generalizable manipulation systems.

\section{Limitations and Future Work}
 
\method requires calibrated RGB-D input at both training and deployment. Depth (sensed or estimated) and camera extrinsics in the robot base frame are needed to lift each pixel into the shared coordinate system that the positional encoding relies on. Removing this requirement, such as by jointly estimating extrinsics or operating on monocular RGB with learned depth priors, is an important direction for future work. Finally, we demonstrate sim-to-real as a proof of concept on a single object category; scaling these findings to large, heterogeneous real-world datasets such as DROID~\cite{khazatsky2024droid} remains an open question.

\bibliographystyle{IEEEtran}
\bibliography{bibliography}

\begin{thebibliography}{10}
\providecommand{\url}[1]{#1}
\csname url@rmstyle\endcsname
\providecommand{\newblock}{\relax}
\providecommand{\bibinfo}[2]{#2}
\providecommand\BIBentrySTDinterwordspacing{\spaceskip=0pt\relax}
\providecommand\BIBentryALTinterwordstretchfactor{4}
\providecommand\BIBentryALTinterwordspacing{\spaceskip=\fontdimen2\font plus
\BIBentryALTinterwordstretchfactor\fontdimen3\font minus
  \fontdimen4\font\relax}
\providecommand\BIBforeignlanguage[2]{{%
\expandafter\ifx\csname l@#1\endcsname\relax
\typeout{** WARNING: IEEEtran.bst: No hyphenation pattern has been}%
\typeout{** loaded for the language `#1'. Using the pattern for}%
\typeout{** the default language instead.}%
\else
\language=\csname l@#1\endcsname
\fi
#2}}

\bibitem{chi2023diffusionpolicy}
C.~Chi, S.~Feng, Y.~Du, Z.~Xu, E.~Cousineau, B.~Burchfiel, and S.~Song,
  ``Diffusion policy: Visuomotor policy learning via action diffusion,'' in
  \emph{Proceedings of Robotics: Science and Systems (RSS)}, 2023.

\bibitem{zhao2023learning}
T.~Z. Zhao, V.~Kumar, S.~Levine, and C.~Finn, ``Learning fine-grained bimanual
  manipulation with low-cost hardware,'' \emph{Robotics: Science and Systems
  (RSS)}, 2023.

\bibitem{Ze2024DP3}
Y.~Ze, G.~Zhang, K.~Zhang, C.~Hu, M.~Wang, and H.~Xu, ``3d diffusion policy:
  Generalizable visuomotor policy learning via simple 3d representations,'' in
  \emph{Proceedings of Robotics: Science and Systems (RSS)}, 2024.

\bibitem{Wang2025articubot}
Y.~Wang, Z.~Wang, M.~Nakura, P.~Bhowal, C.-L. Kuo, Y.-T. Chen, Z.~Erickson, and
  D.~Held, ``Articubot: Learning universal articulated object manipulation
  policy via large scale simulation,'' in \emph{Robotics: Science and Systems
  (RSS)}, 2025.

\bibitem{jiang2026knowyourcamera}
T.~Jiang, J.~Ji, X.~Tan, J.~Fang, A.~Bhattad, V.~Guizilini, and M.~R. Walter,
  ``Do you know where your camera is? {V}iew-invariant policy learning with
  camera conditioning,'' in \emph{IEEE International Conference on Robotics and
  Automation (ICRA)}, 2026.

\bibitem{kleeraven}
D.~Klee, B.~Hu, A.~Cole, H.~Tian, D.~Wang, R.~Platt, and R.~Walters, ``Raven:
  End-to-end equivariant robot learning with rgb cameras,'' in \emph{The
  International Conference on Learning Representations}, 2026.

\bibitem{lin2025depth}
H.~Lin, S.~Chen, J.~H. Liew, D.~Y. Chen, Z.~Li, G.~Shi, J.~Feng, and B.~Kang,
  ``Depth anything 3: Recovering the visual space from any views,'' \emph{arXiv
  preprint arXiv:2511.10647}, 2025.

\bibitem{su2021roformer}
J.~Su, Y.~Lu, S.~Pan, B.~Wen, and Y.~Liu, ``Roformer: Enhanced transformer with
  rotary position embedding,'' 2021.

\bibitem{tian2024vista}
S.~Tian, B.~Wulfe, K.~Sargent, K.~Liu, S.~Zakharov, V.~Guizilini, and J.~Wu,
  ``View-invariant policy learning via zero-shot novel view synthesis,''
  \emph{arXiv}, 2024.

\bibitem{yuan2024learning}
Z.~Yuan, T.~Wei, S.~Cheng, G.~Zhang, Y.~Chen, and H.~Xu, ``Learning to
  manipulate anywhere: A visual generalizable framework for reinforcement
  learning,'' \emph{arXiv preprint arXiv:2407.15815}, 2024.

\bibitem{khazatsky2024droid}
A.~Khazatsky, K.~Pertsch, S.~Nair, A.~Balakrishna, S.~Dasari, S.~Karamcheti,
  S.~Nasiriany, M.~K. Srirama, L.~Y. Chen, K.~Ellis, \emph{et~al.}, ``Droid: A
  large-scale in-the-wild robot manipulation dataset,'' \emph{arXiv preprint
  arXiv:2403.12945}, 2024.

\bibitem{deshpande2026molmobot}
A.~Deshpande, M.~Guru, R.~Hendrix, S.~Jauhri, A.~Eftekhar, R.~Tripathi,
  M.~Argus, J.~Salvador, H.~Fang, M.~Wallingford, \emph{et~al.}, ``Molmob0t:
  Large-scale simulation enables zero-shot manipulation,'' \emph{arXiv preprint
  arXiv:2603.16861}, 2026.

\bibitem{seo2023multi}
Y.~Seo, J.~Kim, S.~James, K.~Lee, J.~Shin, and P.~Abbeel, ``Multi-view masked
  world models for visual robotic manipulation,'' in \emph{International
  Conference on Machine Learning}, 2023, pp. 30\,613--30\,632.

\bibitem{lee2025class}
S.-W. Lee, X.~Kang, B.~Y. Yang, and Y.-L. Kuo, ``Class: Contrastive learning
  via action sequence supervision for robot manipulation,'' in \emph{Conference
  on Robot Learning}.\hskip 1em plus 0.5em minus 0.4em\relax PMLR, 2025, pp.
  4743--4766.

\bibitem{yin2026emergent}
P.~Yin, T.~Westenbroek, Z.~Zhang, J.~Tran, I.~Dagnino, E.~Shilamkar,
  N.~Mbiziwo-Tiapo, S.~Bagaria, X.~Liu, G.~Mullins, A.~Kolobov, and A.~Gupta,
  ``Emergent dexterity via diverse resets and large-scale reinforcement
  learning,'' in \emph{The Fourteenth International Conference on Learning
  Representations}, 2026.

\bibitem{singh2025dextrah}
R.~Singh, A.~Allshire, A.~Handa, N.~Ratliff, and K.~V. Wyk, ``Dextrah-rgb:
  Visuomotor policies to grasp anything with dexterous hands,'' 2025.

\bibitem{ankile2025imitation}
L.~Ankile, A.~Simeonov, I.~Shenfeld, M.~Torne, and P.~Agrawal, ``From imitation
  to refinement-residual rl for precise assembly,'' in \emph{2025 IEEE
  International Conference on Robotics and Automation (ICRA)}.\hskip 1em plus
  0.5em minus 0.4em\relax IEEE, 2025, pp. 01--08.

\bibitem{xue2025opening}
H.~Xue, T.~He, Z.~Wang, Q.~Ben, W.~Xiao, Z.~Luo, X.~Da, F.~Casta{\~n}eda,
  G.~Shi, S.~Sastry, \emph{et~al.}, ``Opening the sim-to-real door for humanoid
  pixel-to-action policy transfer,'' \emph{arXiv preprint arXiv:2512.01061},
  2025.

\bibitem{shridhar2023perceiver}
M.~Shridhar, L.~Manuelli, and D.~Fox, ``Perceiver-actor: A multi-task
  transformer for robotic manipulation,'' in \emph{Conference on Robot
  Learning}.\hskip 1em plus 0.5em minus 0.4em\relax PMLR, 2023, pp. 785--799.

\bibitem{goyal2023rvt}
A.~Goyal, J.~Xu, Y.~Guo, V.~Blukis, Y.-W. Chao, and D.~Fox, ``Rvt: Robotic view
  transformer for 3d object manipulation,'' in \emph{Conference on Robot
  Learning}.\hskip 1em plus 0.5em minus 0.4em\relax PMLR, 2023, pp. 694--710.

\bibitem{goyal2024rvt}
A.~Goyal, V.~Blukis, J.~Xu, Y.~Guo, Y.-W. Chao, and D.~Fox, ``Rvt2: Learning
  precise manipulation from few demonstrations,'' \emph{RSS}, 2024.

\bibitem{yang2026remap}
X.~Yang, R.~Wu, J.~Liu, and X.~Li, ``Remap-dp: Reprojected multi-view aligned
  pointmaps for diffusion policy,'' \emph{arXiv preprint arXiv:2603.14977},
  2026.

\bibitem{ge2025vggt}
S.~Ge, Y.~Zhang, S.~Xie, W.~Zhang, M.~Zhou, and Z.~Wang, ``Vggt-dp:
  Generalizable robot control via vision foundation models,'' \emph{arXiv
  preprint arXiv:2509.18778}, 2025.

\bibitem{wang2025vggt}
J.~Wang, M.~Chen, N.~Karaev, A.~Vedaldi, C.~Rupprecht, and D.~Novotny, ``Vggt:
  Visual geometry grounded transformer,'' in \emph{Proceedings of CVPR}, 2025,
  pp. 5294--5306.

\bibitem{ke20243d}
T.-W. Ke, N.~Gkanatsios, and K.~Fragkiadaki, ``3d diffuser actor: Policy
  diffusion with 3d scene representations,'' \emph{arXiv preprint
  arXiv:2402.10885}, 2024.

\bibitem{gkanatsios20253d}
N.~Gkanatsios, J.~Xu, M.~Bronars, A.~Mousavian, T.-W. Ke, and K.~Fragkiadaki,
  ``3d flowmatch actor: Unified 3d policy for single-and dual-arm
  manipulation,'' \emph{arXiv preprint arXiv:2508.11002}, 2025.

\bibitem{peebles2023scalable}
W.~Peebles and S.~Xie, ``Scalable diffusion models with transformers,'' in
  \emph{Proceedings of the IEEE/CVF international conference on computer
  vision}, 2023, pp. 4195--4205.

\bibitem{alayrac2022flamingo}
J.-B. Alayrac, J.~Donahue, P.~Luc, A.~Miech, I.~Barr, Y.~Hasson, K.~Lenc,
  A.~Mensch, K.~Millican, M.~Reynolds, \emph{et~al.}, ``Flamingo: a visual
  language model for few-shot learning,'' \emph{Advances in neural information
  processing systems}, vol.~35, pp. 23\,716--23\,736, 2022.

\bibitem{jiang2023vima}
Y.~Jiang, A.~Gupta, Z.~Zhang, G.~Wang, Y.~Dou, Y.~Chen, L.~Fei-Fei,
  A.~Anandkumar, Y.~Zhu, and L.~Fan, ``Vima: Robot manipulation with multimodal
  prompts,'' 2023.

\bibitem{nvidia2025groot}
J.~Bjorck, F.~Casta{\~n}eda, N.~Cherniadev, X.~Da, R.~Ding, L.~Fan, Y.~Fang,
  D.~Fox, F.~Hu, S.~Huang, \emph{et~al.}, ``Gr00t n1: An open foundation model
  for generalist humanoid robots,'' \emph{arXiv preprint arXiv:2503.14734},
  2025.

\bibitem{black2024pi_0}
K.~Black, N.~Brown, D.~Driess, A.~Esmail, M.~Equi, C.~Finn, N.~Fusai, L.~Groom,
  K.~Hausman, B.~Ichter, \emph{et~al.}, ``$\pi_0$: A vision-language-action
  flow model for general robot control,'' \emph{arXiv preprint
  arXiv:2410.24164}, 2024.

\bibitem{oquab2023dinov2}
M.~Oquab, T.~Darcet, T.~Moutakanni, H.~Vo, M.~Szafraniec, V.~Khalidov,
  P.~Fernandez, D.~Haziza, F.~Massa, A.~El-Nouby, \emph{et~al.}, ``Dinov2:
  Learning robust visual features without supervision,'' \emph{arXiv preprint
  arXiv:2304.07193}, 2023.

\bibitem{krishna2026ghost}
S.~Krishna, B.~Eisner, H.~Zhan, Y.~Yuan, H.~Zhen, C.~Gan, S.~Tulsiani, and
  D.~Held, ``Ghost: Hierarchical sub-goal policies for generalizing robot
  manipulation,'' in \emph{Robotics: Science and Systems (RSS)}, 2026.

\bibitem{xiang2020sapien}
F.~Xiang, Y.~Qin, K.~Mo, Y.~Xia, H.~Zhu, F.~Liu, M.~Liu, H.~Jiang, Y.~Yuan,
  H.~Wang, \emph{et~al.}, ``Sapien: A simulated part-based interactive
  environment,'' in \emph{Proceedings of the IEEE/CVF conference on computer
  vision and pattern recognition}, 2020, pp. 11\,097--11\,107.

\bibitem{vaswani2017attention}
A.~Vaswani, N.~Shazeer, N.~Parmar, J.~Uszkoreit, L.~Jones, A.~N. Gomez,
  {\L}.~Kaiser, and I.~Polosukhin, ``Attention is all you need,''
  \emph{Advances in neural information processing systems}, vol.~30, 2017.

\bibitem{mittal2025isaaclab}
M.~Mittal, P.~Roth, J.~Tigue, A.~Richard, O.~Zhang, P.~Du, A.~Serrano-Munoz,
  X.~Yao, R.~Zurbr{\"u}gg, N.~Rudin, \emph{et~al.}, ``Isaac lab: A
  gpu-accelerated simulation framework for multi-modal robot learning,''
  \emph{arXiv preprint arXiv:2511.04831}, 2025.

\bibitem{wen2026fastfoundationstereo}
B.~Wen, S.~Dewan, and S.~Birchfield, ``{Fast-FoundationStereo}: Real-time
  zero-shot stereo matching,'' \emph{CVPR}, 2026.

\bibitem{dalal2024manipgen}
M.~Dalal, M.~Liu, W.~Talbott, C.~Chen, D.~Pathak, J.~Zhang, and
  R.~Salakhutdinov, ``Local policies enable zero-shot long-horizon
  manipulation,'' \emph{International Conference of Robotics and Automation},
  2025.

\end{thebibliography}
\clearpage
\appendices

\section{Randomization Details}
 
\subsection{Demonstration Generation}
 
We summarize the initial-state randomization for the cabinet (simulation) and microwave (sim-to-real) opening tasks in Table~\ref{tab:demo_gen_randomization}. Each parameter is sampled independently per trajectory during demonstration generation. In addition, the robot's initial arm joints are sampled such that the robot end-effector is within $0.8$\,m of the object. For both tasks, camera poses are sampled to cover a broad range of over-the-shoulder placements (Table~\ref{tab:camera_randomization}). Each camera pose is sampled independently, and the camera pose is re-sampled if the object handle is not visible at the initial timestep.

\subsection{Visual Domain Randomization}
 
Table~\ref{tab:render_randomization} lists the visual domain randomization applied during IsaacLab rendering for sim-to-real transfer on the microwave task, where we randomize the appearance of the object, robot, table, and background. Backgrounds are drawn uniformly from 485 HDR environment maps from PolyHaven, and object and table textures from 437 material sets from ambientCG. To narrow the sim-to-real gap, we measure the table at our robot workstation and instantiate an analogous table in simulation, randomizing its dimensions about these real-world measurements.
 
For robust sim-to-real transfer, we perturb the camera parameters and depth maps to make the policy robust against such noise in the real world. We perturb the wrist and external cameras by applying a small amount of jitter to the camera focal length, orientation, and translation such that there is a small mismatch between the camera configuration used to render the images and the recorded camera poses. Additionally, we follow~\cite{Wang2025articubot, dalal2024manipgen} and model edge artifacts in depth maps by introducing correlated depth noise using bilinear interpolation on a shifted depth map.
 
\begin{table*}[!t]
\centering
\caption{Visual domain randomization applied in IsaacLab for sim-to-real transfer. Visual appearance, lighting, and camera intrinsics/extrinsics are randomized on top of the demonstration trajectories.}
\label{tab:render_randomization}
\small
\setlength{\tabcolsep}{6pt}
\begin{tabular}{l l}
\toprule
Parameter & Range / Distribution \\
\midrule
\multicolumn{2}{l}{\textit{Object (microwave) appearance}} \\
\quad Material type             & $50\%$ PBR texture / $50\%$ solid color \\
\quad Solid color (per channel) & $\mathcal{U}(0.0, 1.0)$ \\
\quad Texture scale             & $\mathcal{U}(0.7, 5.0)$ \\
\quad Roughness                 & $\mathcal{U}(0.0, 1.0)$ \\
\quad Metallic                  & $\mathcal{U}(0.0, 1.0)$ \\
\quad Specular level            & $\mathcal{U}(0.0, 1.0)$ \\
\midrule
\multicolumn{2}{l}{\textit{Robot appearance}} \\
\quad Arm material color (per channel)   & $\mathcal{U}(0.8, 1.0)$ \\
\quad Gripper base color                 & $(0.78, 0.80, 0.82) + \mathcal{U}(-0.08, 0.08)$ per channel \\
\quad Gripper metallic                   & $\mathcal{U}(0.85, 1.0)$ \\
\quad Gripper roughness                  & $\mathcal{U}(0.10, 0.30)$ \\
\midrule
\multicolumn{2}{l}{\textit{Table}} \\
\quad Length (X) / Width (Y)             & $\mathcal{U}(0.78, 0.92)$\,m / $\mathcal{U}(1.10, 1.35)$\,m \\
\quad Thickness                          & $\mathcal{U}(0.02, 0.08)$\,m \\
\quad Center X offset / Center Y         & $\mathcal{U}(-0.06, 0.06)$\,m / $\mathcal{U}(-0.12, 0.12)$\,m \\
\quad Surface height jitter (Z)          & $\mathcal{U}(-0.025, 0.025)$\,m \\
\quad Yaw                                & $\mathcal{U}(-5^\circ, 5^\circ)$ \\
\quad Surface gray level                 & $\mathcal{U}(0.85, 0.98)$ \\
\midrule
\multicolumn{2}{l}{\textit{Lighting and background}} \\
\quad HDRI environment map               & uniform over map set \\
\quad Dome light intensity               & $\mathcal{U}(100, 3000)$ \\
\quad Dome light rotation (Z)            & $\mathcal{U}(0^\circ, 360^\circ)$ \\
\midrule
\multicolumn{2}{l}{\textit{Camera intrinsics and calibration noise}} \\
\quad Shoulder / wrist focal length      & $\times\,\mathcal{U}(0.98, 1.02)$ / $\times\,\mathcal{U}(0.99, 1.01)$ \\
\quad Wrist mount pitch (down)           & $\mathcal{U}(15^\circ, 25^\circ)$ \\
\quad Wrist mount yaw / roll             & $\mathcal{U}(-3^\circ, 3^\circ)$ each \\
\quad Wrist translation offset           & $\mathcal{U}(0, 0.015)$\,m, random direction \\
\quad Shoulder extrinsic noise (trans / rot) & $\mathcal{N}(0, 0.005\,\text{m})$ / $\mathcal{N}(0, 0.5^\circ)$ \\
\quad Wrist extrinsic noise (trans / rot)    & $\mathcal{N}(0, 0.015\,\text{m})$ / $\mathcal{N}(0, 2.0^\circ)$ \\
\quad Depth edge noise (shoulder / wrist, per frame) & $\sigma = 1.0$\,px / $1.5$\,px, corr.\ length $18$\,px \\
\bottomrule
\end{tabular}
\end{table*}

\begin{table}[!t]
\centering
\caption{Randomization applied during demonstration generation for the Cabinet (sim) and microwave (real) opening tasks. Object pose, scale, and initial robot/object configurations are sampled per trajectory.}
\label{tab:demo_gen_randomization}
\footnotesize
\setlength{\tabcolsep}{4pt}
\begin{tabular}{@{}p{0.34\columnwidth} p{0.60\columnwidth}@{}}
\toprule
Parameter & Range / Distribution \\
\midrule
Object size ratio          & $\mathcal{U}(0.7, 0.9) \times$ base size\\
Object position (X, Y)     & $\mathcal{U}(-0.1, 0.1)$\,m about base, per axis \\
Object yaw                 & base $+\ \mathcal{U}(-30^\circ, 30^\circ)$ \\
Door initial joint angle   & $\mathcal{U}(0, 0.2) \times$ joint range \\
Robot arm initial joints   & per-joint $\mathcal{U}$ within Franka limits, shrunk $20\%$ each side \\
\bottomrule
\end{tabular}
\end{table}
 
\begin{table}[!t]
\centering
\caption{Camera pose randomization for both cabinet and microwave tasks. All parameters are sampled per trajectory.}
\label{tab:camera_randomization}
\footnotesize
\setlength{\tabcolsep}{4pt}
\begin{tabular}{@{}p{0.34\columnwidth} p{0.60\columnwidth}@{}}
\toprule
Parameter & Range / Distribution \\
\midrule
Distance to target  & $\mathcal{U}(0.8, 1.2)\,\text{m} + \mathcal{N}(0, 0.05)\,\text{m}$ \\
Look-at target      & $(0.7, 0.0, 0.4) + \mathcal{N}(0, 0.05)$ per axis \\
Shoulder-cam pitch  & $\mathcal{U}(-20^\circ, 20^\circ)$ \\
Shoulder-cam roll   & $\mathcal{U}(-40^\circ, 0^\circ)$ \\
Shoulder-cam yaw    & $\mathcal{U}(-160^\circ,-110^\circ)$ or $\mathcal{U}(-70^\circ,-20^\circ)$ \\
\bottomrule
\end{tabular}
\end{table}

\begin{figure*}[p]
  \centering
    \includegraphics[width=\textwidth,height=0.88\textheight,keepaspectratio]{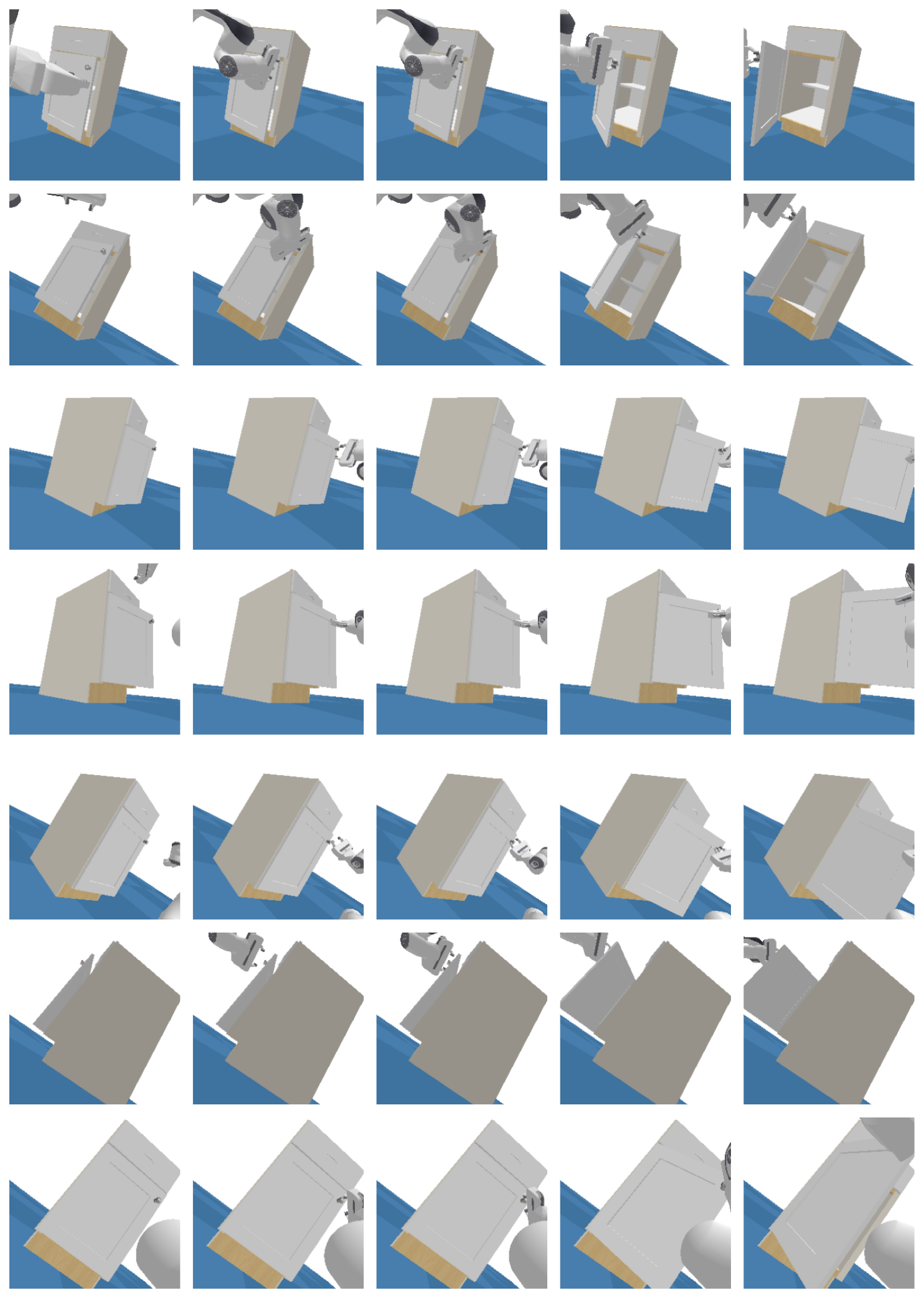}
    \caption{Visualization of the Cabinet task dataset. Each trajectory is rendered from a random camera pose.}
  \label{fig:cabinet_grid}
\end{figure*}
 
\begin{figure*}[p]
  \centering
  \includegraphics[width=\textwidth,height=0.88\textheight,keepaspectratio]{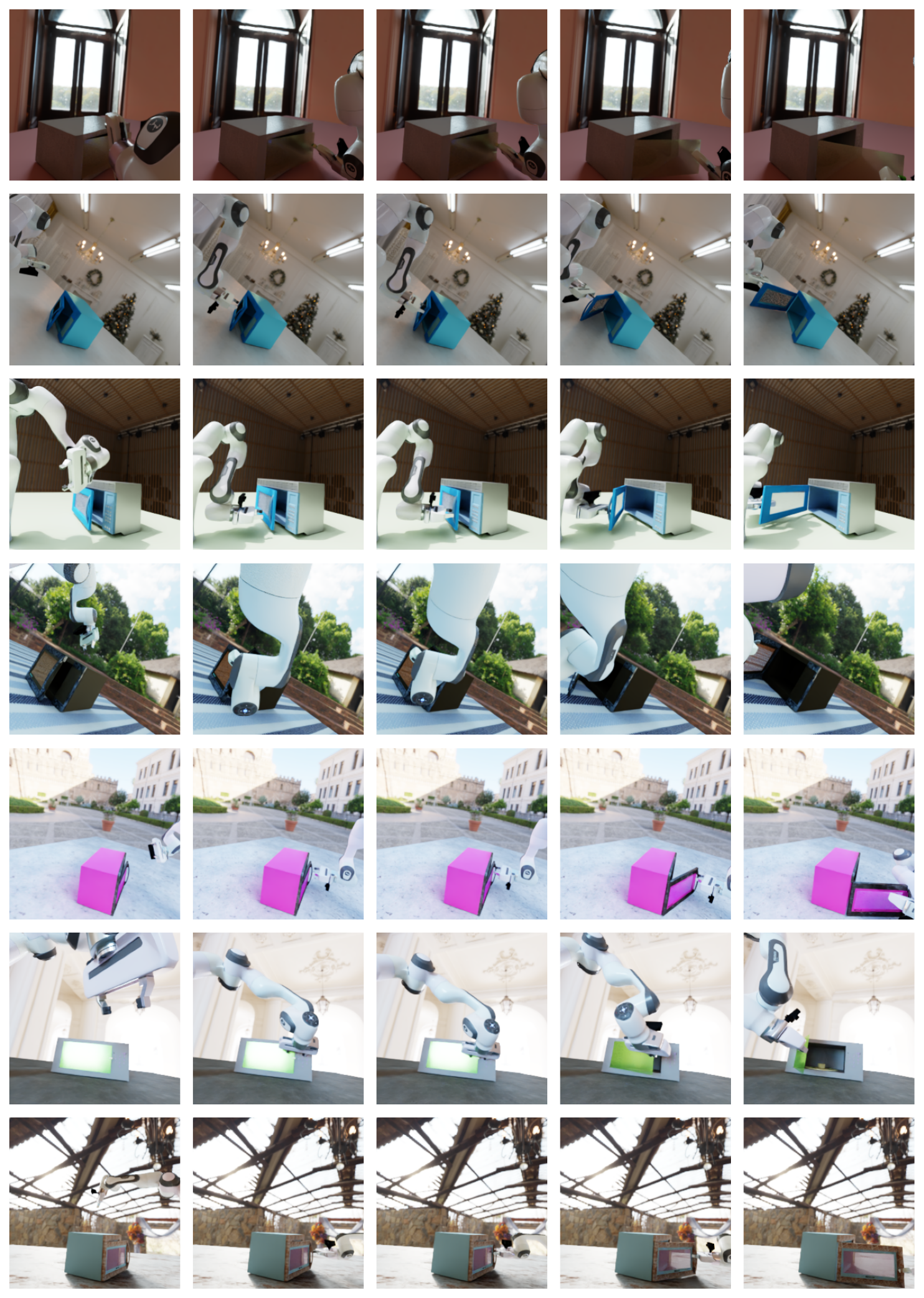}  
  \caption{Visualization of the microwave opening task dataset. Each trajectory is rendered from a random camera pose in IsaacLab.}
  \label{fig:microwave_grid}
\end{figure*}

\end{document}